\documentclass[sigconf, anonymous=false]{acmart}
\usepackage{graphicx}
\usepackage{amsmath}
\usepackage{subfigure}
\usepackage{subcaption}
\usepackage{makecell}
\usepackage{multirow}
\AtBeginDocument{%
  \providecommand\BibTeX{{%
    \normalfont B\kern-0.5em{\scshape i\kern-0.25em b}\kern-0.8em\TeX}}}

\setcopyright{acmlicensed}
\copyrightyear{2018}
\acmYear{2018}
\acmDOI{XXXXXXX.XXXXXXX}

\acmConference[Conference acronym 'XX]{Make sure to enter the correct
  conference title from your rights confirmation emai}{June 03--05,
  2018}{Woodstock, NY}
\acmISBN{978-1-4503-XXXX-X/18/06}

\begin{document}

\title{Compressed models are NOT miniature versions of large models}

\author{Rohit Raj Rai}
\email{rohitraj@iitg.ac.in}
\orcid{1234-5678-9012}

\affiliation{%
  \institution{Indian Institute of Technology Guwahati \country{India}}
  %\city{City}
  %\state{State}
  %\country{Country}
}
\author{Rishant Pal}
\email{rishantpal707@gmail.com}
\orcid{1234-5678-9012}

\affiliation{%
  \institution{Indian Institute of Technology Guwahati \country{India}}
  %\city{City}
  %\state{State}
  %\country{Country}
}
\author{Amit Awekar}
\email{awekar@iitg.ac.in}
\orcid{1234-5678-9012}

\affiliation{%
  \institution{Indian Institute of Technology Guwahati \country{India}}
  %\city{City}
  %\state{State}
  %\country{Country}
}

\renewcommand{\shortauthors}{Rai and Pal, et al.}

\begin{abstract}
 Large neural models are often compressed before deployment. Model compression is necessary for many practical reasons, such as inference latency, memory footprint, and energy consumption. Compressed models are assumed to be miniature versions of corresponding large neural models. However, we question this belief in our work. We compare compressed models with corresponding large neural models using four model characteristics: prediction errors, data representation, data distribution, and vulnerability to adversarial attack. We perform experiments using the BERT-large model and its five compressed versions. For all four model characteristics, compressed models significantly differ from the BERT-large model. Even among compressed models, they differ from each other on all four model characteristics. Apart from the expected loss in model performance, there are major side effects of using compressed models to replace large neural models.
\end{abstract}






\maketitle

\section{Introduction}
%What is model compression
Model compression is the process of compressing a large neural model (LNM) into a smaller model. In the recent past, many model compression techniques such as quantization, pruning, and knowledge distillation have been developed \cite{10.1145/3487045}. Model compression is necessary while deploying LNMs on end-user and edge devices. Such devices have limited memory, computing power, and battery capacity. In addition, LNMs have high inference latency, which leads to low throughput. Compressed models provide a viable option to deploy neural models for applications that require low latency.

\begin{table}[]
\begin{tabular}{|l|c|c|c|c|}
\hline
\multicolumn{1}{|c|}{\textbf{\begin{tabular}[c]{@{}c@{}}Model \\ Name\end{tabular}}} & \textbf{\begin{tabular}[c]{@{}c@{}}Layers\end{tabular}} & \textbf{\begin{tabular}[c]{@{}c@{}}Parameters\\ (Million)\end{tabular}} & \textbf{\begin{tabular}[c]{@{}c@{}}Embedding\\ Dimension\end{tabular}} & \textbf{\begin{tabular}[c]{@{}c@{}}Size \\ (MB)\end{tabular}} \\ \hline
BERT-large                                                                           & 24                                                                   & 340                                                                                  & 1024                                                         & 1340                                                                  \\ \hline
BERT-base                                                                            & 12                                                                   & 110                                                                                  & 768                                                          & 440                                                                   \\ \hline
Distil BERT                                                                          & 6                                                                    & 66                                                                                   & 768                                                          & 268                                                                   \\ \hline
BERT medium                                                                          & 8                                                                    & 41.7                                                                                 & 512                                                          & 167                                                                   \\ \hline
BERT mini                                                                            & 4                                                                    & 11.3                                                                                 & 256                                                          & 45.1                                                                  \\ \hline
Tiny BERT                                                                            & 2                                                                    & 4.4                                                                                  & 128                                                          & 17.7                                                                  \\ \hline
\end{tabular}
\caption{BERT family models used in our experiments}
\label{tab:rel}
\vspace{-6mm}
\end{table}

%Assumption that compressed models is a miniature version of LNM
A compressed model is considered to be a miniature version of the corresponding LNM. Even the naming convention for the compressed models reflects the same belief. For example, various compressed versions of BERT-large models are named Distil-BERT, Tiny-BERT, and BERT-mini. When such compressed models are used in the place of the original LNM, they are expected to behave just like LNM with some reduction in the model performance. Research works on model compression evaluate the behavior of compressed models mostly on just one characteristic: model performance using measures such as precision, recall, and accuracy.

%How to evaluate this assumption?
We evaluate the assumption that compressed models are miniature versions of LNMs. If a compressed model shows high similarity to LNM for most of the model characteristics, then it can be called a miniature version of the LNM. We consider four model characteristics: Prediction errors, Data representation, Data distribution, and Vulnerability to adversarial attacks. For prediction errors, we finetune the six models and compare the set of test points on which model prediction went wrong. For data representation, we compare the $K$ nearest neighbors of a data point across the given models. For data distribution, we use an existing method, Deep Nearest Neighbors \cite{pmlr-v162-sun22d}, to compare the Out-of-Distribution prediction for a challenge dataset. For vulnerability to adversarial attacks, we use the well-known BERT-ATTACK \cite{li-etal-2020-bert-attack} to target all six models.

%Experimental results
We have performed experiments with BERT-large \cite{devlin-etal-2019-bert} as the LNM and its five compressed versions: BERT-base \cite{devlin-etal-2019-bert}, Distil-BERT \cite{sanh2019distilbert}, BERT-medium, BERT-mini, and Tiny-BERT \cite{turc2019}. We used extractive question answering as the finetuning task with the SQUAD2 \cite{rajpurkar2016squad} as the training dataset. For Out-of-Distribution detection, we used the NewsQA \cite{trischler-etal-2017-newsqa} as the challenge dataset. We performed the BERT attack using the IMDB reviews dataset \cite{maas-EtAl:2011:ACL-HLT2011}. For all four model characteristics, we observed that the compressed models differ significantly from the original LNM. Even among the compressed models, the model behavior has very low similarity. We have observed similar results for two other families of models: Bio-BERT \cite{10.1093/bioinformatics/btz682} and T5 \cite{raffel2020exploring}. However, they are not included here due to space limitations. For complete reproducibility, all our code, data, models, and additional experimental results are publicly available on the Web\footnote{Link to be provided in the camera-ready version.}.

\begin{figure*}[t]
    \centering
    \subfigure[Before Fine-tuning]{\includegraphics[height=0.23\textheight]{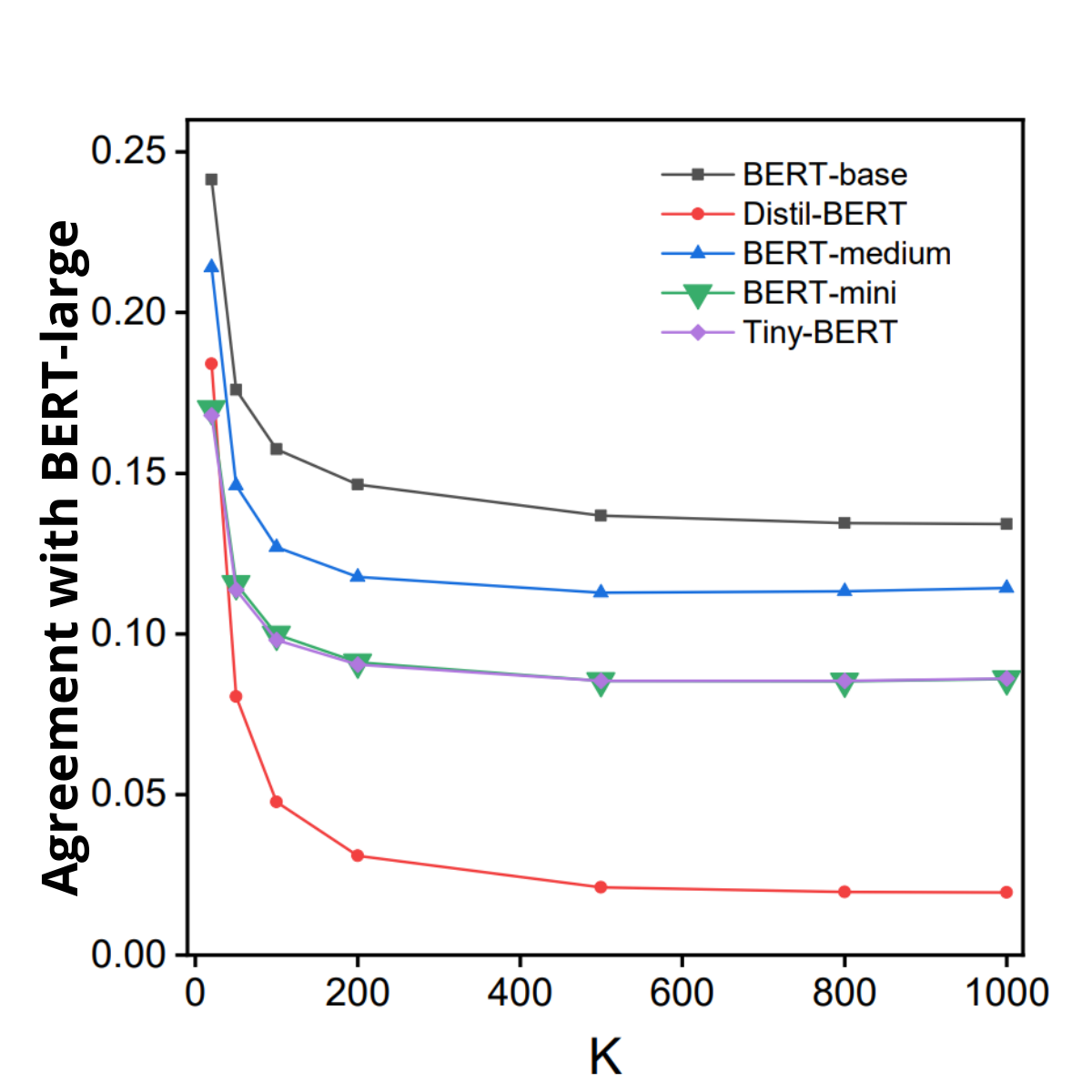}}   
    \subfigure[After Fine-tuning on SQUAD2]{\includegraphics[height=0.23\textheight]{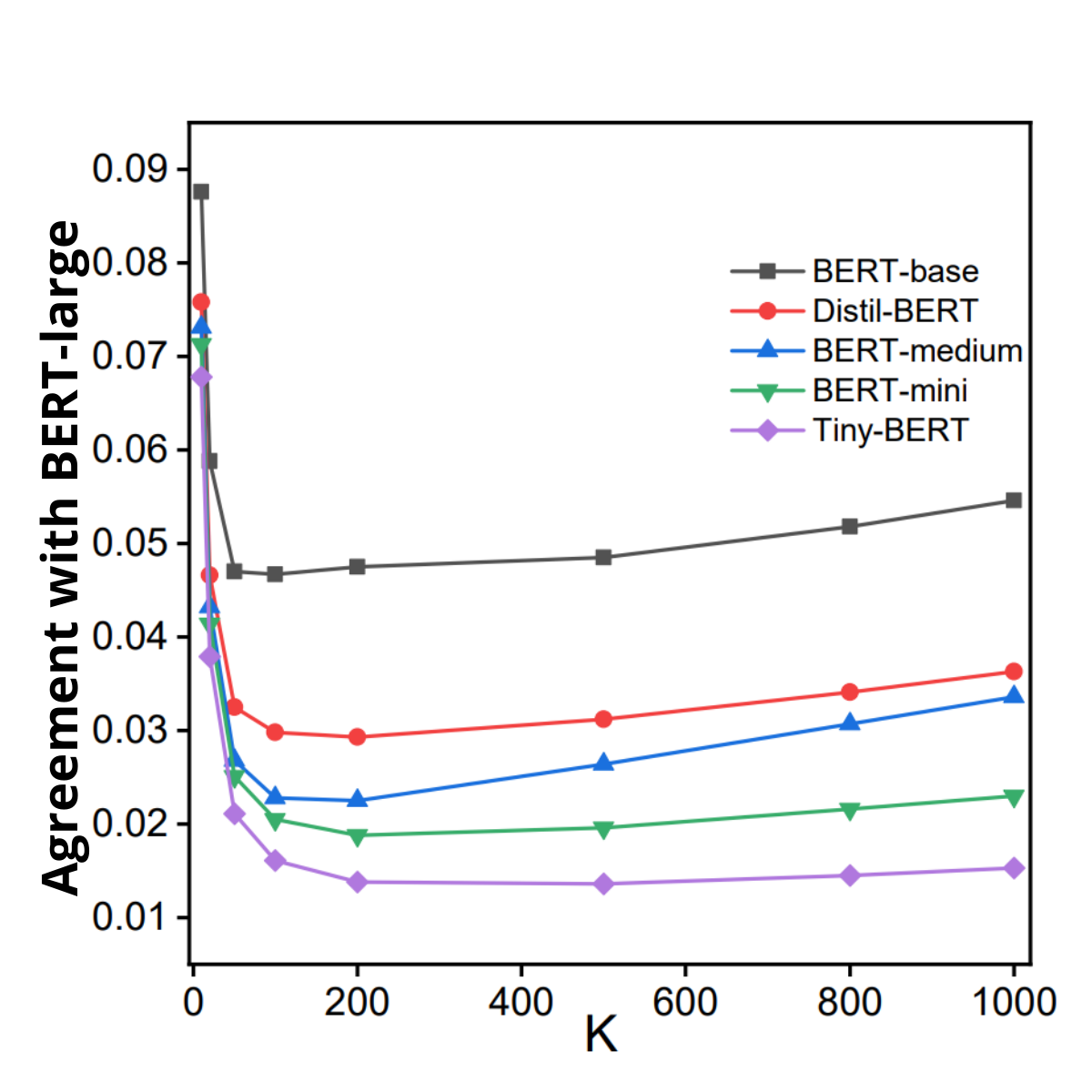}} 
    %\subfigure[Before Fine-tuning]{\includegraphics[width=0.24\textwidth]{data_distribution.pdf}}
    \subfigure[NEWSQA as test dataset]{\includegraphics[height=0.23\textheight]{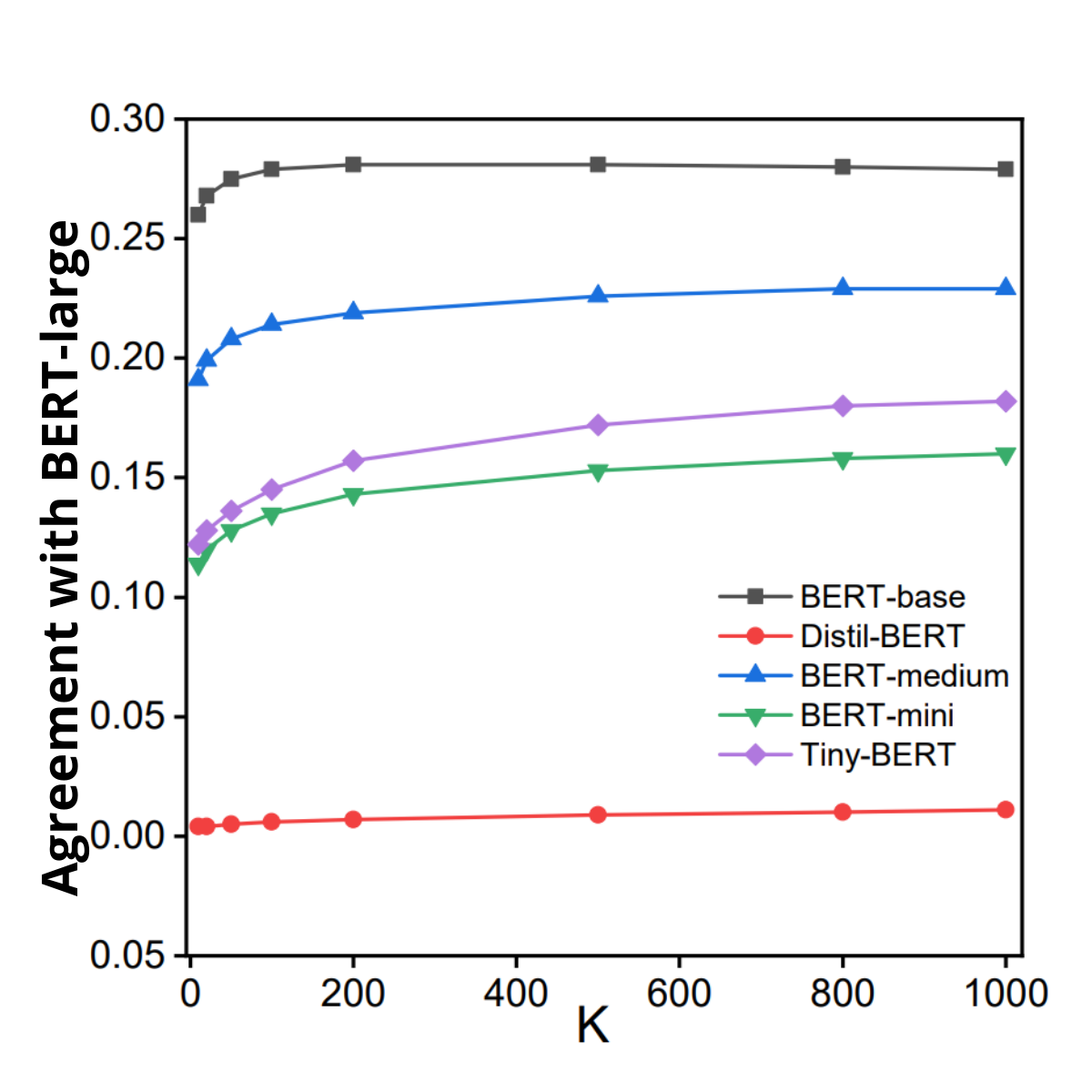}}
    \caption{Variation in Data Representation agreement [(a) and (b)] and Data Distribution agreemtn [(c)] with change in the value of $K$}
    \label{fig:foobar}
\end{figure*}

The primary research contribution of our work is to demonstrate the side effects of using compressed models as a replacement for LNMs. A slight reduction in the model performance is the only anticipated effect of using compressed models. However, our work shows that the reality is far more complex. More careful model behaviour analysis is required before deploying compressed models for real-world applications.

% \usepackage{multirow}
\begin{table*}[]
\begin{tabular}{|c|c|cccccc|}
\hline
\multirow{4}{*}{\textbf{Model $M_i$}} & \multirow{4}{*}{\textbf{Model $M_j$}} & \multicolumn{6}{c|}{\textbf{Model Pair Agreement Expressed as Jaccard Coefficient}}                                                                                                                                                                                                                                  \\ \cline{3-8} 
                                      &                                       & \multicolumn{1}{c|}{\multirow{3}{*}{\textbf{\begin{tabular}[c]{@{}c@{}}Prediction\\ Errors\end{tabular}}}} & \multicolumn{4}{c|}{\textbf{Data Representation}}                                                                                                         & \multirow{3}{*}{\textbf{Data Distribution}} \\ \cline{4-7}
                                      &                                       & \multicolumn{1}{c|}{}                                                                                      & \multicolumn{2}{c|}{\textbf{Before fine tuning}}                            & \multicolumn{2}{c|}{\textbf{After fine tuning}}                             &                                             \\ \cline{4-7}
                                      &                                       & \multicolumn{1}{c|}{}                                                                                      & \multicolumn{1}{c|}{\textbf{Mean}} & \multicolumn{1}{c|}{\textbf{Variance}} & \multicolumn{1}{c|}{\textbf{Mean}} & \multicolumn{1}{c|}{\textbf{Variance}} &                                             \\ \hline
BERT-large                            & BERT-base                             & \multicolumn{1}{c|}{0.528}                                                                                 & \multicolumn{1}{c|}{0.3649}        & \multicolumn{1}{c|}{0.043}             & \multicolumn{1}{c|}{0.0876}        & \multicolumn{1}{c|}{0.0037}            & 0.26                                        \\ \cline{2-8} 
                                      & Distil-BERT                           & \multicolumn{1}{c|}{0.41}                                                                                  & \multicolumn{1}{c|}{0.3652}        & \multicolumn{1}{c|}{0.04}              & \multicolumn{1}{c|}{0.0758}        & \multicolumn{1}{c|}{0.002}             & 0.004                                       \\ \cline{2-8} 
                                      & BERT-medium                           & \multicolumn{1}{c|}{0.421}                                                                                 & \multicolumn{1}{c|}{0.336}         & \multicolumn{1}{c|}{0.05}              & \multicolumn{1}{c|}{0.0731}        & \multicolumn{1}{c|}{0.0024}            & 0.191                                       \\ \cline{2-8} 
                                      & BERT-mini                             & \multicolumn{1}{c|}{0.317}                                                                                 & \multicolumn{1}{c|}{0.266}         & \multicolumn{1}{c|}{0.032}             & \multicolumn{1}{c|}{0.0713}        & \multicolumn{1}{c|}{0.0018}            & 0.114                                       \\ \cline{2-8} 
                                      & Tiny-BERT                             & \multicolumn{1}{c|}{0.22}                                                                                  & \multicolumn{1}{c|}{0.2624}        & \multicolumn{1}{c|}{0.03}              & \multicolumn{1}{c|}{0.0678}        & \multicolumn{1}{c|}{0.0013}            & 0.122                                       \\ \hline
\multirow{4}{*}{BERT-base}            & Distil-BERT                           & \multicolumn{1}{c|}{0.487}                                                                                 & \multicolumn{1}{c|}{0.3995}        & \multicolumn{1}{c|}{0.045}             & \multicolumn{1}{c|}{0.0862}        & \multicolumn{1}{c|}{0.0031}            & 0.004                                       \\ \cline{2-8} 
                                      & BERT-medium                           & \multicolumn{1}{c|}{0.502}                                                                                 & \multicolumn{1}{c|}{0.3879}        & \multicolumn{1}{c|}{0.05}              & \multicolumn{1}{c|}{0.0835}        & \multicolumn{1}{c|}{0.0036}            & 0.264                                       \\ \cline{2-8} 
                                      & BERT-mini                             & \multicolumn{1}{c|}{0.37}                                                                                  & \multicolumn{1}{c|}{0.3145}        & \multicolumn{1}{c|}{0.038}             & \multicolumn{1}{c|}{0.081}         & \multicolumn{1}{c|}{0.0028}            & 0.162                                       \\ \cline{2-8} 
                                      & Tiny-BERT                             & \multicolumn{1}{c|}{0.244}                                                                                 & \multicolumn{1}{c|}{0.3001}        & \multicolumn{1}{c|}{0.035}             & \multicolumn{1}{c|}{0.0761}        & \multicolumn{1}{c|}{0.0021}            & 0.17                                        \\ \hline
\multirow{3}{*}{Distil-BERT}          & BERT-medium                           & \multicolumn{1}{c|}{0.566}                                                                                 & \multicolumn{1}{c|}{0.3941}        & \multicolumn{1}{c|}{0.05}              & \multicolumn{1}{c|}{0.0823}        & \multicolumn{1}{c|}{0.0027}            & 0.004                                       \\ \cline{2-8} 
                                      & BERT-mini                             & \multicolumn{1}{c|}{0.477}                                                                                 & \multicolumn{1}{c|}{0.2885}        & \multicolumn{1}{c|}{0.04}              & \multicolumn{1}{c|}{0.0805}        & \multicolumn{1}{c|}{0.0025}            & 0.008                                       \\ \cline{2-8} 
                                      & Tiny-BERT                             & \multicolumn{1}{c|}{0.297}                                                                                 & \multicolumn{1}{c|}{0.2953}        & \multicolumn{1}{c|}{0.036}             & \multicolumn{1}{c|}{0.0753}        & \multicolumn{1}{c|}{0.0018}            & 0.004                                       \\ \hline
\multirow{2}{*}{BERT-medium}          & BERT-mini                             & \multicolumn{1}{c|}{0.47}                                                                                  & \multicolumn{1}{c|}{0.3466}        & \multicolumn{1}{c|}{0.0361}            & \multicolumn{1}{c|}{0.0795}        & \multicolumn{1}{c|}{0.0026}            & 0.161                                       \\ \cline{2-8} 
                                      & Tiny-BERT                             & \multicolumn{1}{c|}{0.292}                                                                                 & \multicolumn{1}{c|}{0.3257}        & \multicolumn{1}{c|}{0.035}             & \multicolumn{1}{c|}{0.074}         & \multicolumn{1}{c|}{0.002}             & 0.244                                       \\ \hline
Tiny-BERT                             & BERT-mini                             & \multicolumn{1}{c|}{0.487}                                                                                 & \multicolumn{1}{c|}{0.3454}        & \multicolumn{1}{c|}{0.038}             & \multicolumn{1}{c|}{0.0966}        & \multicolumn{1}{c|}{0.004}             & 0.176                                       \\ \hline
\end{tabular}
\caption{Model behaviour comparison for three model characteristics}
\label{tab:1}
%\vspace{-4mm}
\end{table*}

\section{Related Work}
%Model compression methods.
Model compression methods can be broadly partitioned into two categories. First, that reduce the size of the given LNM directly. Second, that train a separate small neural model using the given LNM. Compression methods such as pruning \cite{srinivas2015data} and quantization \cite{pmlr-v139-kim21d} fall in the first category. While the knowledge distillation \cite{hinton2015distilling} method falls in the second paradigm. Pruning identifies the least important parts of the given neural network and deletes them. Pruning can be applied to the edges or neurons in the neural network or the whole layer of the neural network. Quantization does not delete any part of the neural network. Instead, it uses less number of bits to represent weights in the neural network. In the knowledge distillation process, the LNM acts as a teacher, and the target small neural network acts as a student. The student tries to imitate the output of the teacher and, in the process, distills the knowledge of LNM to a small neural network.

%Model characteristics
The behaviour of a neural model can be expressed in terms of its characteristics. Most of the research on Deep Learning and model compression focuses on a single characteristic: model performance. It is the performance of the model measured on a test dataset in terms of measures such as precision, recall, and accuracy. However, other model characteristics also play a crucial role in deciding the appropriate model for the given application. For example, legal compliance in certain domains will require the model to be bias-free as much as possible \cite{sevim2023gender}. In some cases, the explainability of the models is essential. To the best of our knowledge, in the research literature, compressed models are not compared with original LNMs with respect to different model characteristics.

%BERT-large and BERT family of models.
Bidirectional Encoder Representations from Transformers (BERT) \cite{devlin-etal-2019-bert} is a popular deep learning model used for various natural language processing tasks such as classification, question answering, and named entity recognition. It is pre-trained using two tasks: Masked Language Model (MLM) and Next Sentence Prediction (NSP) using the Wikipedia corpus. Two different types of BERT models were initially released: BERT-base and BERT-large. Later on, several other compressed versions of BERT were released by Google in their paper \cite{DBLP:journals/corr/abs-1908-08962}. For our experiments, we have used Distil-BERT \cite{sanh2019distilbert}, BERT-medium, BERT-mini, and Tiny-BERT \cite{turc2019}. These smaller models are created using the idea of knowledge distillation \cite{hinton2015distilling} by using the original BERT as a teacher model. Please refer to Table \ref{tab:rel} for the details of all models used in our experiments.

We have used three datasets: SQuAD2, NewsQA, and IMDBReviews. Stanford Question Answering Dataset (SQUAD) \cite{rajpurkar2016squad} is a reading comprehension dataset. It consists of questions from a set of Wikipedia articles, where the answer to every question denotes a segment of text, or span, from the corresponding reading passage, or there may be no answer to a given question. We have used the version two of the SQuAD dataset (SQuAD2.0) for our experiments\footnote{\url{https://rajpurkar.github.io/SQuAD-explorer/}}. It combines the 100,000 questions in SQuAD1.1 with over 50,000 unanswerable questions. The NewsQA dataset\cite{trischler-etal-2017-newsqa} contains 119,633 natural language questions prepared by crowd workers on 12,744 news articles from CNN/Daily Mail corpus \cite{hermann2015teaching}. Articles in NewsQA are, on average, 6 times longer than the SQuAD2.0 dataset. The IMDb Movie Review Dataset is a well-known dataset in natural language processing used for tasks like sentiment analysis \cite{qaisar2020sentiment}. It consists of 50,000 reviews, labeled as either positive or negative. The train and test partitions each contain 25,000 samples. Each partition contains an equal number of positive and negative reviews. The dataset is available from the hugging face dataset library \cite{maas-EtAl:2011:ACL-HLT2011}. For our experiments, we have used 1,000 samples from the test partition.

\begin{table*}[hbt]
\begin{tabular}{|c|c|c|l|c|c|}
\hline
\textbf{Attack model}      & \textbf{Target model} & \textbf{\begin{tabular}[c]{@{}c@{}}Original \\ accuracy\end{tabular}} & \multicolumn{1}{c|}{\textbf{\begin{tabular}[c]{@{}c@{}}After attack \\ accuracy\end{tabular}}} & \textbf{\begin{tabular}[c]{@{}c@{}}Absolute change\\ in accuracy\end{tabular}} & \textbf{\begin{tabular}[c]{@{}c@{}}\% change\\ in accuracy\end{tabular}} \\ \hline
\multirow{6}{*}{BERT-base} & bert large            & 0.943944                                                              & 0.5635                                                                                         & 0.380444                                                                       & 40.30366                                                                 \\ \cline{2-6} 
                           & bert                  & 0.912913                                                              & 0.22                                                                                           & 0.692913                                                                       & 75.90132                                                                 \\ \cline{2-6} 
                           & distil bert           & 0.921922                                                              & 0.3440                                                                                         & 0.577922                                                                       & 62.68665                                                                 \\ \cline{2-6} 
                           & medium bert           & 0.906907                                                              & 0.3140                                                                                         & 0.592907                                                                       & 65.37682                                                                 \\ \cline{2-6} 
                           & mini bert             & 0.888889                                                              & 0.2460                                                                                         & 0.642889                                                                       & 72.325                                                                   \\ \cline{2-6} 
                           & tiny bert             & 0.850851                                                              & 0.11                                                                                           & 0.740851                                                                       & 87.07177                                                                 \\ \hline
\end{tabular}
\caption{Model behaviour comparison for BERT-ATTACK}
\label{tab:attck_3}
%\vspace{-4mm}
\end{table*}

\section{Prediction Errors}
The first model characteristic we considered was the errors made by the model. Two models can be considered similar if they make similar mistakes on the test data. We finetuned the six models (BERT-large and five compressed models) for the extractive question-answering task using the SQUAD2 dataset. We tested them using the test partition of the same dataset. A model is considered to have made an error on a test data point if its answer fails to match the ground truth answer given in the dataset. For each model $M_i$, we have the set $E_i$ that represents the set of test data points for which $M_i$ failed to produce the correct answer. Then, the similarity between the two models  $M_i$ and $M_j$ can be computed as the Jaccard coefficient of sets $E_i$ and $E_j$. The Jaccard coefficient of any two sets is the size of their intersection divided by the size of their union.

Please refer to Table ~\ref{tab:1}. Each row of the table represents a model pair to compare. Each column represents similarity measurements for three model characteristics. The first five rows represent the similarity between the LNM and five compressed models. The rest of the rows represent the similarity among the compressed models. Please refer to the column corresponding to the Prediction Errors. We can observe that all the compressed models have low similarity scores with the LNM. In other words, the errors made by the compressed models are not similar to those made by the LNM. In addition, we also observed that the error set of the LNM is not a subset of the error set of any compressed model. 

\section{Data Representation}
BERT is an encoder-only transformer-based model. For a sequence of tokens received as input, it produces a vector representation for each token along with a CLS token that summarises the whole input. The CLS token can be considered as the representation of the data generated by any BERT family model. The CLS token for the same input across two models cannot be directly compared. 

We can compare the most similar data points for the given data point using the cosine similarity. Given a data point $P$, let $KNN_i(P)$ represent the set of $K$ nearest neighbors for $P$ using the model $M_i$. Then the agreement between two models $M_i$ and $M_j$ for the data representation of $P$ can be computed as the Jaccard coefficient of sets $KNN_i(P)$ and $KNN_j(P)$. For the whole dataset with $n$ data points, the data representation agreement between $M_i$ and $M_j$ can be computed as the mean across all data points.

\[\sum_{d=1}^{n} \frac{|KNN_i(P_d) \cap KNN_j(P_d)|}{|KNN_i(P_d) \cup KNN_j(P_d)|} * \frac{1}{n} \]

Please refer to Table ~\ref{tab:1} and columns under the heading Data Representation. There are four columns. We can compute the data representation agreement between the models in two ways: before finetuning and after finetuning. We compute the mean and variance of agreement between the models for both scenarios. To finetune the models, we use the SQUAD2 dataset for the extractive question answering task. First, the mean value of the Jaccard coefficient between the BERT-large and all the compressed models is quite low. After finetuning, the value goes down further. The variance is negligible as compared to the mean value. Results for both the mean and variance indicate that compressed models have very low agreement with the LNM for data representation. We can see a similar trend when we compute data representation agreement between various compressed models. In order to compute the $K$-nearest neighbors efficiently, we have used the \texttt{FAISS} library \footnote{\url{https://github.com/facebookresearch/faiss}} developed by Facebook AI Research team. 

The results in Table~\ref{tab:1} are computed with $K$ set to 10. We also ensured that our results were not sensitive to the choice of the value of $K$. Please refer to Figures 1(a) and 1(b). The X-axis shows the variation in the value of $K$. The Y-axis shows the mean Jaccard coefficient. We have plotted the variation in data representation agreement between BERT-large and the five compressed models. We can observe that the mean Jaccard coefficient is low even with a considerable variation in the value of $K$. The observation holds for both scenarios: before and after finetuning.

 \section{Data Distribution}
Test data provided to the model after deployment might not match with the training data on which the model was trained. Such data is called Out-of-Distribution (OOD) with respect to the training data and the model. OOD detection methods classify the test data point into in-distribution or OOD. Deep Nearest Neighbors (DNN) \cite{pmlr-v162-sun22d} is one of the recent methods specifically designed for detecting OOD test data for Deep Learning models. It is based on computing the distance between the test data point and the training dataset. To perform the OOD detection task, the DNN method computes the distance to the $K$th nearest neighbor in the training dataset for each test data point. The test data point is classified as OOD if the distance is above a threshold.

We use the DNN method to compare the OOD predictions of various models. Consider two models $M_i$ and $M_j$ with common test dataset $D_c$. $M_i$ predicts that the set $O_i$ ($O_i\subset D_c$) is the OOD component of $D_C$. Similarly, $M_j$ predicts that $O_j$ ($O_j\subset D_c$) is the OOD component of $D_C$. Then, the data distribution agreement between $M_i$ and $M_j$ can be computed as the Jaccard coefficient of sets $O_i$ and $O_j$.

Please refer to Table ~\ref{tab:1} and column under the heading Data Distribution. We have used the SQUAD2 dataset to finetune all the models. We have used the NEWSQA dataset as the test dataset. We can observe that the agreement between BERT-large and any compressed models is low. The same is the case when we compute agreement between the compressed models. The DNN method requires a value of $K$ as the parameter. The results in Table ~\ref{tab:1} are calculated with $K$ set to 10. We wanted to ensure that our results are not sensitive to the value of $K$. Please refer to Figure 1(c). The X-axis represents the variation in the value of $K$. The Y-axis shows the Jaccard coefficient for data distribution agreement of compressed models with BERT-large. We can observe that the Jaccard coefficient does not change significantly with the value of $K$.

\section{Vulnerability to Adversarial Attacks}
Two models can be considered similar if they have similar vulnerability to adversarial attacks. There are several adversarial attacks designed for BERT-based models \cite{morris2020textattack} \cite{keller2021bert} \cite{jha2023codeattack}. One of the popular attacks is the BERT-ATTACK \cite{li-etal-2020-bert-attack}. This attack considers two models: attack and target. Given a test data point, the attack first finds the top $K$ important words from the data point that influence the model's output. Then, for each important word, the attack model is used to find semantically similar replacement words. Then, the modified input is provided to the target model. If the prediction of the target model changes, the attack is considered successful.

We used the BERT-base model as the attack model and experimented with all six models as the target models. We have used the sentiment classification task as the example task and the IMDB reviews dataset as the training dataset. Each data point contains a review of the movie, classified as positive or negative. We have set the value of $K$ to 3 for our experiments.

Please refer to Table ~\ref{tab:attck_3}. First, we computed the accuracy of each target model on the test dataset after finetuning on the IMDB reviews dataset. We can observe that the accuracy of the compressed models is slightly lower than that of the BERT-large model. However, when we attack these models using the BERT-base, there is significant variation in the success of the attack. Performance of BERT-large degrades only by 40\%. Nevertheless, the performance loss is far more significant for the compressed models. The smallest model, Tiny-BERT, has the highest performance degradation. This result indicates that compressed models are more vulnerable to adversarial attacks than the corresponding LNMs.

\section{Conclusion and Future Work}
We have compared BERT-lage with its five compressed versions. We analyzed four characteristics of these models: Prediction errors, Data representation, Data distribution, and Vulnerability to adversarial attacks. The compressed models differed significantly from the BERT-large model on all four characteristics. Our work shows that we cannot assume the compressed models to be miniature versions of LNMs. While deploying the compressed models as a replacement for LNMs, we should be aware of the complex changes in the model characteristics. One-dimensional focus on the model performance is not enough to predict the behaviour of compressed models. Our work can be extended by developing novel model compression methods that try to retain multiple characteristics of the original LNM.

\bibliographystyle{ACM-Reference-Format}
\bibliography{sample-base}

% \appendix




\end{document}